\documentclass[a4paper,11pt,oneside,fleqn]{article}
\usepackage[utf8]{inputenc}
\usepackage[ngerman,english]{babel}
\usepackage[T1]{fontenc}
\usepackage{lmodern}
\usepackage{mathtools}
\usepackage{hyperref}
\usepackage{enumerate}
\usepackage{amscd}
\usepackage{graphicx}
\usepackage{caption}
\usepackage{subcaption}
\usepackage{extarrows}
\usepackage{faktor}
\usepackage[vcentermath]{youngtab}
\usepackage{mathtools}
\usepackage{color}
\usepackage{colortbl}
\usepackage{makeidx}
\usepackage[all]{xy}
\usepackage{amsmath,amssymb,amsthm,amsfonts}
\usepackage{paralist}
\usepackage{enumitem}
\usepackage{tikz,pgf}
\usepackage{tikz-cd}
\usetikzlibrary{matrix, arrows}

\usepackage{adjustbox}
\usepackage{algpseudocode}
\usepackage{algorithm}
\usepackage{multirow}

{\theoremstyle{definition}

}

\hypersetup{colorlinks, linkcolor=blue, citecolor=blue, urlcolor=blue}

\newcommand{\todo}[1][\null]{\ensuremath{\clubsuit}}

\title{Model-agnostic machine learning of conservation laws from data}
\author{Shivam Arora$^\ddag$, Alex Bihlo$^{\ddag}$, Rüdiger Brecht$^{\flat}$ and Pavel Holba$^{\dag}$
\\~\\
\small{$^{\ddag}$Department of Mathematics and Statistics, Memorial University of Newfoundland,}\\
\small{$\phantom{^{\ddag}}$~St.\ John's (NL) A1C 5S7, Canada}
\\~\\
\small{$^{\flat}$Department of Mathematics, University of Hamburg}\\
\small{$\phantom{^{\flat}}$~Hamburg, Germany}
\\~\\
\small{$^{\dag}$Mathematical Institute, Silesian University in Opava}\\
\small{$\phantom{^{\dag}}$~Opava, Czech Republic}
\\~\\
\small E-mails: sarora17@mun.ca, abihlo@mun.ca, ruediger.brecht@uni-hamburg.de, pavel.holba@math.slu.cz
}

\begin{document}

\maketitle

\vspace{12mm}\par\noindent\hspace*{10mm}\parbox{140mm}{\small
We present a machine learning based method for learning first integrals of systems of ordinary differential equations from given trajectory data. The method is model-agnostic in that it does not require explicit knowledge of the underlying system of differential equations that generated the trajectories. As a by-product, once the first integrals have been learned, also the system of differential equations will be known. We illustrate our method by considering several classical problems from the mathematical sciences.
\par}\vspace{7mm}

\section{Introduction}

Conservation laws play an important role in the mathematical sciences. They provide crucial physical constraints and significant information about structural symmetries  to real-world systems~\cite{olve93a}, and are used extensively in numerical methods for differential equations, see e.g.~\cite{mcla99a,quis96a,wan17a}. 

While both the analytical and numerical investigation of conservation laws have reached a certain maturity in the past several decades, recently machine learning has also been proposed as a means to find conservation laws, see e.g.~\cite{ha21a, liu22a, liu21a}. A main appeal of machine learning is that it allows one to work with raw data, making it particularly relevant for the solution of inverse problems. Indeed, machine learning has been used extensively for solving inverse problems, and notably also to identify differential equations from data, see~\cite{karn21a, rais19a} for some examples. 

The inverse problem of conservation laws, i.e.\ the problem of determining which systems of differential equations admit prescribed conservation laws, has received relatively little attention in the literature, see~\cite{popo20a} for a review and some general results. In this paper we will provide a machine learning approach to the inverse problem on conservation laws, by proposing a method that allows one to identify the conservation laws inherent in given trajectory data. 

More specifically, in this paper we propose a model-agnostic method based on machine learning to learn conservation laws from data. We refer to this method as \textit{model-agnostic} as it does not require the knowledge of an underlying system of differential equations, but rather can work with given trajectory data only. As a by-product, once the conservation laws inherent in the given data have been successfully identified, our method will have also learned the system of differential equations that generated the given trajectory data. It will then also be possible to simulate new trajectories using the learned conservation laws.

This approach has the following two advantages over training neural networks to directly learn trajectories or to just identify the underlying system of differential equations directly.

Firstly, both conservation laws and the associated system of differential equations are being learned. Conservation laws provide insights into the physics and underlying symmetries of the system of differential equations. They thus provide important additional information beyond the mere knowledge of the given system of differential equations itself.

Secondly, the model is trained to learn the underlying dynamics of the system rather than a pure non-linear approximation of trajectory data, which is prone to over-fitting and missing the underlying physics, see~\cite{grey19a}. Knowing the underlying system of differential equations also allows generating new trajectory data from the learned dynamics, thereby providing important generalization capabilities. 

The further organization of this paper is as follows. In the subsequent Section~\ref{sec:RelatedWork} we give an overview of recent work on conservation laws relevant for the present consideration. Section~\ref{sec:Method} introduces our method for learning conservation laws from data and, as a by-product, identifying the associated differential equations underlying the given data. Section~\ref{sec:Applications} provides the numerical results for identifying conservation laws for several important physical systems. The final section~\ref{sec:Conclusions} provides a short summary and discusses potential future directions of research in this field.

\section{Related work}\label{sec:RelatedWork}

Conservation laws have been studied extensively analytically in the past, with such celebrated results as Noether's theorem relating symmetries of Lagrangian systems to conservation laws via a constructive formula~\cite{olve93a}. More generally, conservation laws of differential equations can be found constructively using conservation law characteristics (or multipliers), which can be computed by solving a suitable linear system of determining partial differential equations~\cite{olve93a,blum10a}. This relation between conservation laws and conservation law characteristics is at the heart of the multiplier method for finding conservative discretization schemes to systems of differential equations~\cite{wan17a}. 

Machine learning for problems related to conservation laws is a relatively recent field. Here, a key result is the development of Hamiltonian neural networks~\cite{grey19a}, where a neural network is used to learn/model the Hamiltonian function of a Hamiltonian system. The loss function for this problem are the Hamiltonian equations of motion, which are approximated using trajectory data. Owing to the form of the problem, Hamiltonian neural networks numerically conserve the learned Hamiltonian. Learning conservation laws from trajectory data was the subject of~\cite{ha21a,liu21a}, where a new variance decreasing loss and methods of dimensionality reduction have been used, respectively. In~\cite{liu22a} the authors use the defining equations of the considered systems of differential equations to learn their conservation laws.

Our method has flavours of the works \cite{ha21a,liu22a,liu21a} in that we do use trajectory data for training but end up learning the conservation laws inherent in this data by identifying the differential equation responsible for this data.

\section{Method}\label{sec:Method}

In this section we detail our method for identifying conservation laws of physical systems from trajectory data. Before delving into the details of our method, we briefly present a high-level overview of our approach.

Consider a given unknown system of ordinary differential equations $\dot{\mathbf{x}}=\mathbf{f}(\mathbf{x})$ that admits a single conserved quantity $I$. Our goal is to learn $I$ from known trajectory data, that has been generated using the given system of differential equations.

It was shown in~\cite{mcla99a} that the system of differential equations
$$\dot{\mathbf{x}}=\mathbf{f}(\mathbf{x}) \qquad\text{ can be rewritten as }\qquad \dot{\mathbf{x}}=\epsilon(\mathbf{x})\nabla I(\mathbf{x}),$$
where $\epsilon$ is an anti-symmetric matrix. We then train a neural network on the trajectory data of the system of differential equations to learn both the anti-symmetric matrix $\epsilon$ and the conserved quantity $I$. Once both $\epsilon$ and $I$ have been learned, not only have we learned the conservation law $I$ such that $\dot I = 0$, but the entire system of differential equations as well. It turns out that the skew-gradient form $\dot{\mathbf{x}}=\epsilon(\mathbf{x})\nabla I(\mathbf{x})$ can also be generalized to the case of multiple conserved quantities, and thus an analogous approach can  be used to learn multiple conservation laws simultaneously or iteratively.

In the following, we work exclusively with data stemming from systems of ordinary differential equations and will thus first review the associated terminology of conservation laws, and the associated skew-gradient forms for such systems in the following Section~\ref{sec:first_integrals}. In Section \ref{sec:dis_grad} we then discuss how this skew-gradient form has to be discretized to be applicable to numerical trajectory data. This provides the key for our method to learning first integrals from data, which will be introduced in Section~\ref{sec:ClLearningDiscreteGradient}.

\subsection{First integrals of ordinary differential equations}\label{sec:first_integrals}

We consider the system of ordinary differential equation (ODEs)
\begin{align}\label{eq:ode}
\begin{split}
&\dot{\mathbf{x}}=\mathbf{f}(\mathbf{x})    
\\
&\mathbf{x}(t_0)=\mathbf{x}_0,
\end{split}
\end{align}
where $t\in T\subset \mathbb{R}$, and $\mathbf{x}(t)=(x_1(t),\dots,x_d(t))\in X \subset \mathbb{R}^d$. If $\mathbf{f}\in C^{p-1}(T\times X\to\mathbb{R}^d)$ is a $(p-1)$-times continuously differentiable function over the domain $T\times X$ then system~\eqref{eq:ode} admits a unique solution $\mathbf{x}\in C^p(T\to X)$ in a neighbourhood of $(t_0, \mathbf{x}_0)\in T\times X$.

Conservation laws of systems of ordinary differential equations are typically referred to as \textit{constants of motion} or \textit{first integrals}. In the following, we will refer to them as first integrals.  Mathematically, a first integral is a function $I\in C^1(T\times X\to\mathbb{R})$ such that
\begin{equation}\label{eq:CLzero}
\mathrm{D}_tI(t,\mathbf{x})|_{\dot{\mathbf{x}}-\mathbf{f}(\mathbf{x})= \mathbf{0}} = \mathbf{0}
\end{equation}
holds for any $t\in T$ and any $C^p$ solution $\mathbf{x}$ of system~\eqref{eq:ode}, where $\mathrm{D}_t$ is the total derivative with respect to $t$. That is, $I(t_0, \mathbf{x}_0) = I(t,\mathbf{x})$, and $I$ is constant on $t$ along solutions of  system~\eqref{eq:ode}. We refer to solutions of system~\eqref{eq:ode} as trajectories.

Note that in the following we will allow for system~\eqref{eq:ode} to also include arbitrary parameters $\boldsymbol{\gamma}\in\mathbb{R}^s$, i.e.\ we will consider models of the form $\dot{\mathbf{x}}=\mathbf{f}(\mathbf{x}; \boldsymbol{\gamma})$, and we assume that for each fixed parameter vector $\boldsymbol{\gamma}$ the associated system of differential equations satisfies the above conditions such that a unique solution to that system exists. Also, while first integrals may depend on the time $t$, such as for the damped harmonic oscillator~\cite{liu22a}, in the following we will consider only the case of time-independent first integrals, i.e.\ $I=I(\mathbf{x})$.

If system~\eqref{eq:ode} admits $n$ first integrals $I^i$ with $0<i\leqslant n$, whose gradients are linearly independent, then one can write system \eqref{eq:ode} in the form
\[
 \mathbf{\dot x} = \epsilon(\mathbf{x})\nabla I^1\nabla I^2\cdots\nabla I^n,
\]
or, component-wise, using Einstein's summation rule,
\begin{equation}\label{eq:SkewGradientForm}
 \dot x_i = \epsilon_{ij_1j_2\dots j_n}\partial_{j_1}I^1\partial_{j_2}I^2\cdots\partial_{j_n}I^n,\qquad i=1,\dots, d,
\end{equation}
with each $j_k=1,\dots,d$, $k\in \{1,\dots,n\}$, and where $\epsilon$ is a totally anti-symmetric $(n+1)$-tensor. See~\cite{mcla99a} for further discussions on the skew-gradient form~\eqref{eq:SkewGradientForm}. Two prominent special cases arise from the general skew-gradient form~\eqref{eq:SkewGradientForm}. 

If $n=1$ then this form reduces to
\[
 \dot x_i = \epsilon_{ij}\partial_j I
\]
which for the special case of $\epsilon_{ij}$ being the two-dimensional Levi-Civita symbol and $I$ being the Hamiltonian is the general form of a canonical Hamiltonian system~\cite{olve93a}. 

If $n=2$ then the skew-gradient from~\eqref{eq:SkewGradientForm} reads
\begin{equation}\label{eq:NambuForm}
 \dot x_i = \epsilon_{ijk}\partial_j I^1 \partial_k I^2
\end{equation}
which was first suggested by Nambu as a generalization to Hamiltonian dynamics~\cite{namb73a}. Two prominent examples for Nambu systems are the Euler equation for rigid rotations~\cite{namb73a} and the conservative Lorenz-1963 model~\cite{nevi94a}. We will consider the latter in Section~\ref{sec:Applications}.

\subsection{Discrete gradient methods}\label{sec:dis_grad}

The skew-gradient form~\eqref{eq:SkewGradientForm} is at the heart of the \textit{discrete gradient method}~\cite{mcla99a,quis96a,nort15a}, which is a general purpose method for constructing conservative discretization schemes for systems of ordinary differential equations.

The discrete gradient method requires consistent choices for discretizing the skew-gradient tensor $\epsilon(\mathbf{x})$ and the gradients $\nabla I^i$, $i=1,\dots,n$. Moreover, the gradients have to be chosen in such a manner as to guarantee that the resulting discretization indeed numerically preserves $I^i$. 

Consider the case where $n=1$, i.e.\ a single first integral exists. Then it was show in~\cite{quis96a} that for a one-step method for system~\eqref{eq:ode} of the form
\begin{equation}\label{eq:odeDiscretized}
\frac{\mathbf{x}'-\mathbf{x}}{\Delta t} = \mathbf{\tilde f}(\mathbf{x},\mathbf{x}',\Delta t)
\end{equation}
the following is a discrete gradient of $I$,
\begin{equation}\label{eq:DiscreteGradientConsistency}
\bar i(\mathbf{x},\mathbf{x}')\cdot (\mathbf{x}'-\mathbf{x}) = I(\mathbf{x}') - I(\mathbf{x}),\qquad \bar i(\mathbf{x},\mathbf{x}) = \nabla I(\mathbf{x}),
\end{equation}
and that if the discretization~\eqref{eq:odeDiscretized} is chosen as
\begin{equation}\label{eq:OdeDiscreteGradientSingle}
\frac{\mathbf{x}'-\mathbf{x}}{\Delta t} = \bar \epsilon(\mathbf{x},\mathbf{x}',\Delta t) \bar i(\mathbf{x},\mathbf{x}'),
\end{equation}
then~\eqref{eq:OdeDiscreteGradientSingle} indeed numerically preserves $I$ as long as $\bar \epsilon(\mathbf{x},\mathbf{x}',\Delta t)$ is a consistent approximation to the anti-symmetric matrix $\epsilon(\mathbf{x})$, i.e.\ as long as $\lim_{\Delta t\to 0}\bar \epsilon(\mathbf{x},\mathbf{x}',\Delta t)=\epsilon(\mathbf{x})$. 

Analogously, for general $n$, the one-step method
\begin{equation}\label{eq:OdeDiscreteGradientMultiple}
\frac{x'_i-x_i}{\Delta t} = \bar \epsilon_{ij_1j_2\dots j_n} \bar i^1_{j_1}\bar i^2_{j_2}\cdots\bar i^n_{j_n},
\end{equation}
will preserve the first integrals $I^1,\dots, I^n$ numerically provided that $\bar \epsilon$ is consistent with the anti-symmetric $(n+1)$-tensor $\epsilon$ and $\bar i^k$ are discrete gradients of $I^k$, $k=1,\dots,n$.

Various formulas for discrete gradients exist that satisfy the consistency condition~\eqref{eq:DiscreteGradientConsistency}, see~\cite{mcla99a}. In the following, we use the simple expression
\begin{equation}\label{eq:DiscreteGradientForm}
\bar i(\mathbf{x},\mathbf{x}') = \begin{pmatrix}
\frac{I(x_1',x_2,\dots,x_d)-I(x_1,x_2,\dots,x_d)}{x'_1-x_1}\\
\frac{I(x_1',x_2',\dots,x_d)-I(x_1',x_2,\dots,x_d)}{x'_2-x_2}\\
\vdots\\
\frac{I(x_1',x_2',\dots,x_d')-I(x_1',x_2',\dots,x_{d-1}', x_d)}{x'_d-x_d},
\end{pmatrix},
\end{equation}
which does satisfy~\eqref{eq:DiscreteGradientConsistency}.

We should like to note here that there is a self-consistency among skew-gradient forms. In particular, if system~\eqref{eq:ode} possesses $n$ first integrals then it can be written both as
\begin{subequations}\label{eq:SelfConsistency}
\begin{equation}
\dot x_i = \epsilon_{ij_1j_2\dots j_n}\partial_{j_1}I^1\partial_{j_2}I^2\cdots\partial_{j_n}I^n,
\end{equation}
and as
\begin{equation}
\dot x_i = \tilde\epsilon_{ij_1j_2\dots j_{n-1}}\partial_{j_1}I^1\partial_{j_2}I^2\cdots\partial_{j_{n-1}}I^{n-1},
\end{equation}
where the $n$-tensor $\tilde \epsilon$ is totally anti-symmetric and defined as.
\begin{equation}\label{eq:SelfConsistencyTensor}
\tilde\epsilon_{ij_1j_2\dots j_{n-1}} = \epsilon_{ij_1j_2\dots j_n}\partial_{j_n}I^n.
\end{equation}
\end{subequations}
As a particular example for the self-consistency condition~\eqref{eq:SelfConsistency}, consider the case when $n=2$. System~\eqref{eq:NambuForm} can then be written as
\begin{equation}\label{eq:SelfConsistencyLossPenaltyForm}
 \dot x_i = \epsilon_{ijk}\partial_j I^1 \partial_k I^2 = \tilde\epsilon_{ij}\partial_j I^1 = \hat \epsilon_{ik}\partial_k I^2,
\end{equation}
where
\[
\tilde\epsilon_{ij} = \epsilon_{ijk}\partial_k I^2,\quad \hat \epsilon_{ik} = \epsilon_{ijk}\partial_j I^1.
\]

In the following, we will make extensive use of the discrete gradient method.

\subsection{Learning first integrals using discrete gradient methods} \label{sec:ClLearningDiscreteGradient}

We assume that we have collected trajectory data of the form $(\mathbf{x}^0, \mathbf{x}^1, \dots, \mathbf{x}^N)$ for a physical system stemming from an unknown system of ordinary differential equations, where superscripts denote the time step. For the sake of simplicity, we assume that this trajectory data is sampled at a uniform time step $\Delta t$, although this is not necessary for our method as long as the (variable) sampling time step is known.

Assuming this system has a single first integral, then it can be written as
\[
\mathbf{\dot x} = \epsilon(\mathbf{x})\nabla I(\mathbf{x}),
\]
for some unknown anti-symmetric matrix $\epsilon(\mathbf{x})$ and first integral $I(\mathbf{x})$. A discrete gradient formulation for this equation is given by~\eqref{eq:OdeDiscreteGradientSingle}. Since the trajectory data is given, the task is to identify the unknown anti-symmetric matrix and first integral.

In this work we use a standard feedforward neural network with weights $\boldsymbol{\theta}$ for this identification task. Specifically, our neural network accepts as input pairs of consecutive trajectory time steps $(\mathbf{x}^k,\mathbf{x}^{k+1})$, $k=0,\dots, N-1$ and outputs the associated anti-symmetric matrix $\bar \epsilon_{\boldsymbol{\theta}}(\mathbf{x}^{k}, \mathbf{x}^{k+1},\Delta t)$ and first integral $I_{\boldsymbol{\theta}}(\mathbf{x}^k)$ by minimizing the loss function
\begin{equation}\label{eq:LossFunctionSingleCL}
\mathcal L_{\boldsymbol{\theta}} = \frac{1}{N}\sum_{k=0}^{N-1}\left(\frac{\mathbf{x}^{k+1}-\mathbf{x}^k}{\Delta t}-\bar\epsilon_{\boldsymbol{\theta}}(\mathbf{x}^{k}, \mathbf{x}^{k+1},\Delta t)\bar i_{\boldsymbol{\theta}}(\mathbf{x}^{k}, \mathbf{x}^{k+1})\right)^2,
\end{equation}
where $\bar i_{\boldsymbol{\theta}}(\mathbf{x}^{k}, \mathbf{x}^{k+1})$ denotes the evaluation of the discrete gradient~\eqref{eq:DiscreteGradientForm} using $I_{\boldsymbol{\theta}}$ obtained from the neural network as the first integral $I$.




Assuming that the unknown system of differential equations has more than one first integral, there are several computationally feasible ways for determining them.
\begin{description}
\item[Method 1.] If the number of first integrals was known to be $n$, then one could try to directly minimize the loss analogous to~\eqref{eq:LossFunctionSingleCL} derived from using Eq.~\eqref{eq:OdeDiscreteGradientMultiple}. A main downside of this approach, besides requiring to know $n$, is that the anti-symmetric (n+1)-tensor $\epsilon$ is increasingly difficult to learn for larger $n$ or $d$.

\item[Method 2.] Rather than learning all first integrals and the anti-symmetric (n+1)-tensor $\epsilon$ at once, one can use the self-consistency among skew-gradient forms~\eqref{eq:SelfConsistency}. Here, one neural network is learned after another, where the dimension of the anti-symmetric tensor is increased step by step, see Algorithm~\ref{alg:learning2}.
An advantage of this method is that the number of first integrals does not have to be known beforehand. A downside of this approach is that again large anti-symmetric tensors have to be defined, which we in practice found to slow down learning.

\item[Method 3.] Learn the possible first integrals one by one using~\eqref{eq:SelfConsistencyLossPenaltyForm}, see Algorithm \ref{alg:learning}. The difference to the previous approach is that the anti-symmetric tensor does not change shape. This way the complexity to learn the anti-symmetric tensor and first integral stays the same. The downside of this approach is that the penalty increases the complexity of the loss function at each step. 
\end{description}
%
%



We note that we tried each of these methods to identify the conservation laws and found the results are qualitatively similar. Thus, we will use the third method in the remainder of this work, because it is computationally more efficient.

\begin{algorithm}
\caption{The algorithm to learn the first integrals of an unknown system of differential equations corresponding to Method 3.}\label{alg:learning}
\begin{description} 
\item[Start]
 Train a neural network to learn an anti-symmetric matrix $ \bar\epsilon_{\boldsymbol{\theta_1}}$ and the first integral $I_{\boldsymbol{\theta_1}}$ using the loss~\eqref{eq:LossFunctionSingleCL}.
\item[For] $\ell=2,\ldots,d$~\\
   Create a new neural network learning $I_{\boldsymbol{\theta_\ell}}$ and anti-symmetric matrix $\bar \epsilon_{\boldsymbol{\theta_{\ell}}}$ penalizing it to not recover the already learned anti-symmetric matrices $\bar \epsilon_{\boldsymbol{\theta_1}},\ldots,\bar \epsilon_{\boldsymbol{\theta_{\ell-1}}}$ and first integrals $I_{\boldsymbol{\theta_1}}, \ldots, I_{\boldsymbol{\theta_{\ell-1}}}$. To do so, we follow the ideas of~\cite{liu22a} and define the loss function of the second neural network as
\begin{align*}
\mathcal L_{\boldsymbol{\theta_{\ell}}} &= \frac{1}{N}\sum_{k=0}^{N-1}\bigg[\left(\frac{\mathbf{x}^{k+1}-\mathbf{x}^k}{\Delta t}-\bar\epsilon_{\boldsymbol{\theta_{\ell}}}(\mathbf{x}^{k}, \mathbf{x}^{k+1},\Delta t)\bar i_{\boldsymbol{\theta_{\ell}}}(\mathbf{x}^{k}, \mathbf{x}^{k+1})\right)^2+\\
&\sum_{m=1}^\ell
\alpha \left(\bar i_{\boldsymbol{\theta_m}}(\mathbf{x}^{k}, \mathbf{x}^{k+1})\bar i_{\boldsymbol{\theta_\ell}}(\mathbf{x}^{k}, \mathbf{x}^{k+1})\right)^2\bigg],
\end{align*}
    where the second term, the gradient penalty with penalization constant $\alpha\in\mathbb{R}$, should prevent the new neural network from learning $I_{\boldsymbol{\theta_1}},\ldots,I_{\boldsymbol{\theta_{\ell-1}}}$ again.
~\\~\\
\textbf{If} Explained ratio (See Eq.~\eqref{eq:explainedRatio}) of $I_{\boldsymbol{\theta_\ell}}$ is $<0.01$
$\longrightarrow$ \textbf{ stop}.
\end{description}
\end{algorithm}

\begin{algorithm}
\caption{The algorithm to learn the first integrals of an unknown system of differential equations corresponding to Method 2.}\label{alg:learning2}
\begin{description} 
\item[Start]
 Train a neural network to learn an anti-symmetric  matrix $\bar \epsilon_{\boldsymbol{\theta_1}}$ and the first integral $I_{\boldsymbol{\theta_1}}$ using the loss~\eqref{eq:LossFunctionSingleCL}.
 
 \item[For] $\ell=2,\ldots,d$~\\
Create a new neural network learning $I_{\boldsymbol{\theta_\ell}}$ and anti-symmetric $(\ell+1)$-tensor $\bar \epsilon_{\boldsymbol{\theta}_\ell}$
and with weights $\boldsymbol{\theta}_\ell$ to minimize
    \[
\mathcal L_{\boldsymbol{\theta_{\ell}}} =  \frac{1}{N}\sum_{k=0}^{N-1}\left(
\bar \epsilon_{\boldsymbol{\theta_{\ell-1}}}(\mathbf{x}^k,\mathbf{x}^{k+1},\Delta t) -
 \bar\epsilon_{\boldsymbol{\theta_\ell}}(\mathbf{x}^k,\mathbf{x}^{k+1},\Delta t)\bar i_{\boldsymbol{\theta}_\ell}(\mathbf{x}^k,\mathbf{x}^{k+1})
\right)^2,
\]

\textbf{If} Explained ratio of $I_{\boldsymbol{\theta_\ell}}$ is $<0.01$
$\longrightarrow$ \textbf{ stop}.
\end{description}
\end{algorithm}
\subsubsection{Remark:}\label{rm:funofCLisCL}
A main issue in the case of multiple first integrals is that any function of first integrals is again a first integral. This can make it potentially difficult to identify the most elementary, or the most canonical, form of a set of first integrals. Also, and more importantly, it becomes potentially challenging to identify the minimal set of first integrals that are functionally independent. We will address this issue in Section~\ref{sec:EvaluationMetrics}.

\subsection{Neural network architecture and training}

We use a standard feedforward neural network in this work and report the number of layers and units used for each example below. We use the sigmoid linear unit as activation function except for the output layers where a linear activation function is used. 

The input to the network is a pair of $d$-dimensional data point $(\mathbf{x}^k, \mathbf{x}^{k+1})$ on the sampled trajectory. If the system under consideration admits some model parameters $\boldsymbol{\gamma}$, and if the trajectory data is sampled for different parameter values, then the neural network will also have to accept these model parameters as input. We illustrate this below for the case of the harmonic oscillator, where $\boldsymbol{\gamma}=(m, k)$, the mass and spring constant of the oscillator, for the conservative Lorenz model, where $\boldsymbol{\gamma}=(\sigma, \rho)$, which are related to the Prandtl and Rayleigh numbers, respectively~\cite{lore63a}, and for the point-vortex equations where $\boldsymbol{\gamma}=(\Gamma_1,\dots,\Gamma_p)$ are the vortex strengths of the $p$ point vortices. 

The output of the network depends on which of the above three variations of our method is chosen. If all first integrals are learned at once (assuming the total number of first integrals is known), then the network will output an $n$-dimensional vector of first integrals and an anti-symmetric (n+1)-tensor. Similarly, if we aim to learn the first integrals in a sequential manner, then the output of the neural network for the first first integral will be a single scalar quantity and an appropriately shaped anti-symmetric matrix. For the second first integral the neural network will then have to output another scalar quantity and either an anti-symmetric 3-tensor (using the second method) or another anti-symmetric matrix (using the third method). Additional first integrals are then learned analogously with additional neural networks.

All models have been trained in \texttt{TensorFlow} 2.11 using a single NVIDIA RTX 8000 GPU. 

\subsection{Evaluation metrics}\label{sec:EvaluationMetrics}

In Remark \ref{rm:funofCLisCL} we mentioned that the representation of a set of first integrals is not unique as any function of first integrals is again a first integral. Additionally, owing to the skew-gradient form~\eqref{eq:SkewGradientForm} any multiplication of each first integral can be offset by an associate re-scaling of the skew-symmetric tensor $\epsilon$. Thus, it may be challenging to find the equations for the predicted first integrals in their simplest form. While this is still possible for the known first integrals in the examples presented below, we also introduce the following general purpose metrics to assess the success in learning (functionally independent) first integrals.

Since we learn both the first integrals and the entire system of differential equations from data, we aim to quantify our success in both tasks by:
\begin{itemize}\itemsep=0ex
    \item Computing how well the first integrals are conserved, see Eq. \eqref{eq:CLzero}. That is, we compute the relative error in conservation $(I_{\boldsymbol{\theta}}(\mathbf{x}^k)-I_{\boldsymbol{\theta}}(\mathbf{x}^0))/I_{\boldsymbol{\theta}}(\mathbf{x}^0)$, $k=0,\ldots,N$ for new trajectories $(\mathbf{x}^0,\ldots,\mathbf{x}^N)$.
    \item Using the learned system of differential equations to simulate a new trajectory $(\mathbf{x}^0_{\boldsymbol{\theta}},\dots, \mathbf{x}^N_{\boldsymbol{\theta}})$ and then compare it with the numerical reference trajectory. 
\end{itemize}

To identify the correct number of functionally independent first integrals we adapt the strategy proposed in~\cite{liu22a}. Having computed the $k>1$ first integrals we form the $(Nd)\times k$ matrix $A$ defined by
\begin{equation}\label{eq:matrix_A}
A=\begin{pmatrix}
    \partial_1I^1(\mathbf{x}^0) & \partial_1I^2(\mathbf{x}^0) & \dots & \partial_1I^k(\mathbf{x}^0) \\
    \vdots & \vdots & \vdots & \vdots \\
    \partial_1I^1(\mathbf{x}^N) & \partial_1I^2(\mathbf{x}^N) & \dots & \partial_1I^k(\mathbf{x}^N) \\
    \partial_2I^1(\mathbf{x}^0) & \partial_2I^2(\mathbf{x}^0) & \dots & \partial_2I^k(\mathbf{x}^0) \\
    \vdots & \vdots & \vdots & \vdots \\
    \partial_2I^1(\mathbf{x}^N) & \partial_2I^2(\mathbf{x}^N) & \dots & \partial_2I^k(\mathbf{x}^N) \\
    \vdots & \vdots & \vdots & \vdots \\
    \partial_dI^1(\mathbf{x}^0) & \partial_dI^2(\mathbf{x}^0) &
    \dots & \partial_dI^k(\mathbf{x}^0)\\
    \vdots & \vdots & \vdots & \vdots \\
    \partial_dI^1(\mathbf{x}^N) & \partial_dI^2(\mathbf{x}^N) & \dots & \partial_dI^k(\mathbf{x}^N)
\end{pmatrix},
\end{equation}
where $N\gg k$ is the number of points along one trajectory. We then compute the singular value decomposition of the matrix $A$ and determine the rank of this matrix by the number of non-vanishing singular values. As argued in~\cite{liu22a}, if $N\gg k$ then it is exponentially unlikely that this method will under-sample the true manifold dimensionality, and thus this method will yield the correct rank, and hence the correct number of functionally independent first integrals. This will be verified explicitly in Section~\ref{sec:Applications}. We follow~\cite{liu22a} and treat components of the singular values $\{\sigma_1,\dots, \sigma_k\}$ as vanishing if the explained fraction of the total variance given by 
\begin{equation}\label{eq:explainedRatio}
    \sigma_i^2/\sum_j\sigma_j^2, i=1,\dots,k
\end{equation}
falls below $10^{-2}$. We refer to Eq.~\eqref{eq:explainedRatio} as the \textbf{explained ratio}.

\section{Application to models from the mathematical sciences}\label{sec:Applications}

In this section we illustrate our method with several examples from the mathematical sciences. By default we use $\alpha=0.1$ for the penalty loss.
\begin{table}[!ht]
\begin{adjustbox}{width=\columnwidth,center}
\begin{tabular}
{lp{1.8cm}||p{2.2cm}|p{2.75cm}|p{2.5cm}|p{2.7cm}|p{2.8cm}}
&  & Harmonic oscillator & H\'enon--Heiles model & Lorenz--1963 system & 3D Lotka--Volterra & Point vortex equations \\ \cline{2-7}\noalign{\vskip\doublerulesep\vskip-\arrayrulewidth} \cline{2-7} 
\multirow{6}{*}[-6em]{\rotatebox{90}{ODE}}  
& \textit{Dimension} & 2  & 4 & 3  & 3  & 4  \\ \cline{2-7}
& $\#$ \textit{Cls} & 1  & 1 & 2  & 2  & 4  \\ \cline{2-7}
& \textit{Variable domain}  & $q \in [-1,1]$, $p\in [-1,1]$  & $q_i\in[-0.5, 0.5]$, $p_i\in[-0.1,0.1]$  & $x, y\in [-2,2]$, $z\in[0,1]$           & $x,y,z\in[0.1,2]$ & $x_i\in[0.5,1]$, $y_i\in[-1,-0.5]$ \\ \cline{2-7}
& $T$ & 1  & 10 & 2  & 20  & 150  \\ \cline{2-7}
& $\Delta t$  & 0.01 & 0.01 & 0.01 & 0.01 & 0.05 \\ \cline{2-7}
& \textit{No.\ of trajectories} & 200 & 500 & 100 & 200 & 65 \\ \cline{2-7} 
& \textit{Integration scheme} & symplectic Euler & symplectic Euler & trapezoidal & conservative & symplectic Euler \\ \cline{2-7}
& \textit{Model parameters} & $m\in[0.1,1]$, $k\in[0.1,1]$ & none & $\sigma\in[0.1, 4]$, $\rho\in[0.1,2]$  & none & $\Gamma_1 \in [0.1, 5]$, $\Gamma_2 \in [0.1, 5]$  \\ \cline{2-7}\noalign{\vskip\doublerulesep\vskip-\arrayrulewidth}\cline{2-7}
& & & & & & \\
\multirow{ 4}{*}[0.5em]{\rotatebox{90}{NN}}
& \textit{Layers} & 4 & 6 & 6 & 6 & 6 \\ \cline{2-7}
& \textit{Units} & 40 & 40 & 40 & 40 & 40 \\ \cline{2-7}
& \textit{Epochs} & 1000 & 10000 & 1000 & 2000 & 1000           \end{tabular}
\end{adjustbox}
\caption{In this table we summarize the details for the numerical generation of the trajectories and for the neural network setup. Here, $\#~Cls$ stands for the number of conserved quantities of the given system.  }
\label{Table:parameters}
\end{table}

In Table \ref{Table:parameters} we give the details for the setup of each of the examples. For each example, we sample the initial conditions uniformly in the given domain. Then, using a suitable numerical integrator we integrate the trajectory from $t=0$ to $t=T$ using a time-step of $\Delta t$. For most examples we deliberately choose numerical schemes that are not conservative. We argue that this is a reasonable choice as trajectories sampled from real-world phenomena would never exhibit exactly conservative behaviour due to noise in the data acquisition process. As such, we should like to test here the robustness of our algorithms for data that has only approximately conservative properties. Our default method for all Hamiltonian systems is the symplectic Euler method which does not preserve the Hamiltonian exactly but has no systematic trend in the numerical Hamiltonian~\cite{haie06a}. For the Lorenz--1963 model we choose the trapezoidal method which exhibits a drift in the conserved quantities. For the Lotka--Volterra equations we choose the fully conservative integrator proposed in~\cite{wan17a} to showcase that the error in results obtainable from our method does not appear to be critically dependent on whether or not the underlying data is exactly conservative; in other words, the results for the Lotka--Volterra equations are not qualitatively or quantitatively better than those obtained for the other examples where non-conservative methods were used.  

The architecture for the neural networks has been found using try and error although we experimentally found the method to be rather forgiving provided enough hidden layers and units per layer are available. The number of epochs was chosen to train the network until the loss stagnated at a low level. No regularlization had to be used to train the neural networks.

Once the neural network learned the anti-symmetric tensor and first integrals, they can be used to construct the right hand side of the underlying dynamical systems, and we can then compute new trajectories for each system of differential equations. This is being done using the discrete gradient method~\eqref{eq:OdeDiscreteGradientSingle}, which is a natural choice since the right hand side being learned is automatically in the skew-gradient form being used by this method.

Each of the following systems of differential equations with $n$ equations can have at most $n$ conserved quantities (see \cite{goldstein1980classical}), meaning that $n$ neural networks have to be trained if the total number of first integrals is not known beforehand. The actual number of first integrals found is being determined using the numerical rank computation based on the matrix $A$, see Eq. \eqref{eq:matrix_A}.

\subsection{Harmonic oscillator}

The harmonic oscillator is a Hamiltonian system of the form
\begin{align}\label{eq:hos}
\begin{split}
    \dot{q} = \frac{p}{m},\quad
    \dot{p} = -kq,
\end{split}    
\end{align}
where $m$ is the mass of the oscillator and $k$ is the associated spring constant. The Hamiltonian, which is a first integral, is
\begin{equation}
    I^1 =\frac{1}{2}\left( \frac{p^2}{m}+kq^2 \right).
\end{equation}
Let $\mathbf{x}=(q,p)$ and $\epsilon=\begin{pmatrix}
0&1\\-1&0
\end{pmatrix}$
be the canonical Poisson tensor. We can then rewrite \eqref{eq:hos} in the skew-gradient form
\begin{equation}
    \mathbf{\dot{x}}=\epsilon \nabla I^1,
\end{equation}
and the goal of the neural network is to identify both $\epsilon$ and $I^1$.

\begin{figure}[!ht]
\centering
     \begin{subfigure}[b]{0.32\textwidth}
         \centering
         \includegraphics[width=\textwidth]{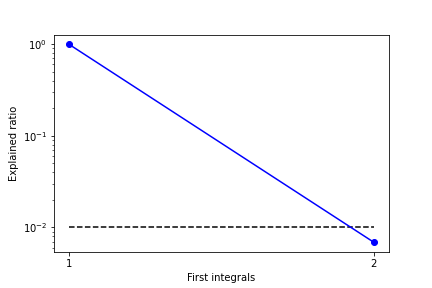}
         \label{sfig:harmonic_rank}
     \end{subfigure}
    \hfill
    \begin{subfigure}[b]{0.32\textwidth}
         \centering
         \includegraphics[width=\textwidth]{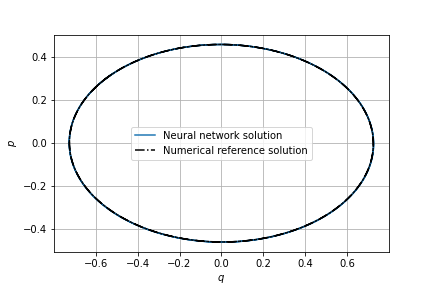}
         \label{sfig:harmonic_h1_simu}
     \end{subfigure}
     \hfill
     \begin{subfigure}[b]{0.32\textwidth}
         \centering
         \includegraphics[width=\textwidth]{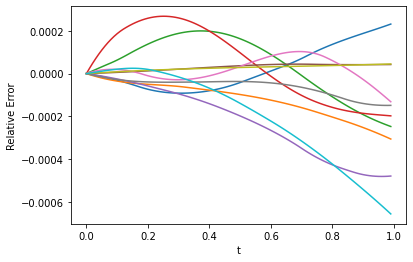}
         \label{sfig:harmonic_h1-h10}
     \end{subfigure}
    \begin{subfigure}[b]{\textwidth}
         \centering
         \includegraphics[width=\textwidth]{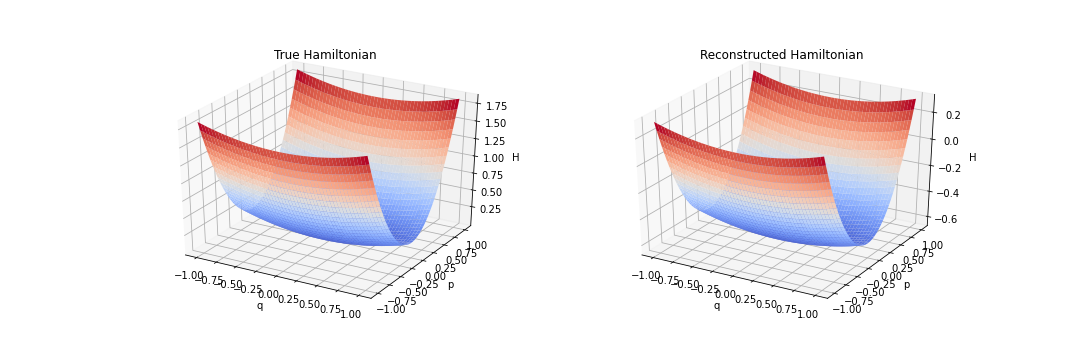}
         \label{sfig:harmonic_h1}
     \end{subfigure}
     \caption{Numerical results for the harmonic oscillator. \textit{Top row (left):} Number of first integrals identified using the explained ratio based on the non-vanishing singular values of matrix $B$. Numerical threshold (dashed line) is set to $0.01$ following~\cite{liu22a}.
     \textit{Top row (center):} Simulation of a trajectory using the learned anti-symmetric matrix $\bar\epsilon_{\boldsymbol{\theta}}$ and first integral $I_{\boldsymbol{\theta}}$. \textit{Top row (right):} Relative conservation error $(I^1_{\boldsymbol{\theta}}(\mathbf{x})-I^1_{\boldsymbol{\theta}}(\mathbf{x}^0))/I^1_{\boldsymbol{\theta}}(\mathbf{x}^0)$ for ten randomly generated trajectories following the learned dynamics. \textit{Bottom row:} True Hamiltonian and shape of the learned Hamiltonian for $k=0.49$ and $m=0.81$.}
    \label{fig:hos_nn_conservation}
\end{figure}

As there are two degrees of freedom for this problem, we trained two neural networks and used the matrix $A$ from Eq.~\eqref{eq:matrix_A} to identify that there exists only one first integral, see Fig.~\ref{fig:hos_nn_conservation}. For evaluation purposes, we randomly sampled a harmonic oscillator with $k=0.49$ and $m=0.81$. We note that this is a harmonic oscillator that the neural network has never encountered during training, and hence illustrates how well the neural network has learned to understand the overall physics of this class of harmonic oscillators. Fig.~\ref{fig:hos_nn_conservation} illustrates that the neural network has learned the anti-symmetric matrix $\epsilon$ and first integral $I^1$ well enough to be able to accurately reproduce the phase-space diagram for the associated harmonic oscillator. In Fig.~\ref{fig:hos_nn_conservation} we also observe that $I_{\boldsymbol{\theta}}$ is conserved to the order of $\mathcal{O}(10^{-3})$ for randomly generated trajectories using the learned anti-symmetric matrix and first integral. Lastly, in Fig.~\ref{fig:hos_nn_conservation} we show the surface of the learned first integral and verify that it closely resembles the true Hamiltonian for the given problem. 

We have also experimented with a variety of other parameter values $m$ and $k$ and verified that the above results hold over the entire range of values $m$ and $k$. This illustrates that our neural networks have indeed learned the overall dynamics and first integral of the entire class of harmonic oscillators.

\subsection{H\'{e}non--Heiles model}

The H\'{e}non--Heiles model 
\begin{align}\label{eq:HenonHeiles}
\begin{split}
    \dot q_1  = p_1, \quad
    \dot q_2  = p_2, \quad
    \dot p_1  = -q_1 - 2q_1q_2, \quad
    \dot p_2  = -q_2 - q_1^2 + q_2^2, 
\end{split}
\end{align}
is a Hamiltonian system with Hamiltonian function
\[
I^1(q_1, q_2, p_1, p_2) = \frac12(q_1^2+q_2^2 + p_1^2 + p_2^2) + q_1^2q_2 - \frac13 q_2^3.
\]
Presented system has applications in astronomy and was studied extensively to answer the question whether a second invariant exists for this model, which is not the case~\cite{haie06a,heno64a}. 

\begin{figure}[!ht]
\centering
     \begin{subfigure}[b]{0.32\textwidth}
         \centering
         \includegraphics[width=\textwidth]{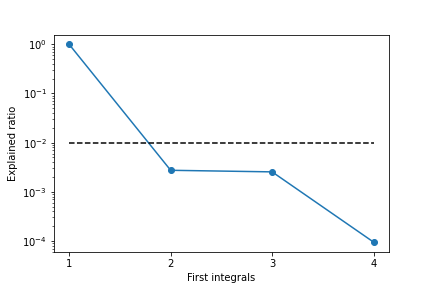}
         \label{sfig:hh_rank}
     \end{subfigure}
    \hfill
    \begin{subfigure}[b]{0.32\textwidth}
         \centering
         \includegraphics[width=\textwidth]{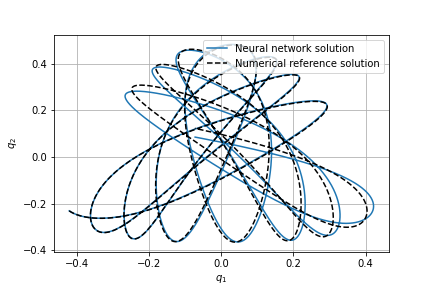}
         \label{sfig:hh_h1_simu}
     \end{subfigure}
     \hfill
     \begin{subfigure}[b]{0.32\textwidth}
         \centering
         \includegraphics[width=\textwidth]{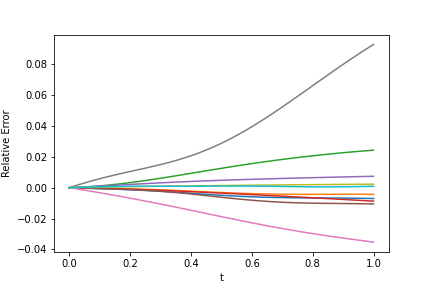}
         \label{sfig:hh_conservation}
     \end{subfigure}
    \begin{subfigure}[b]{\textwidth}
         \centering
         \includegraphics[width=\textwidth]{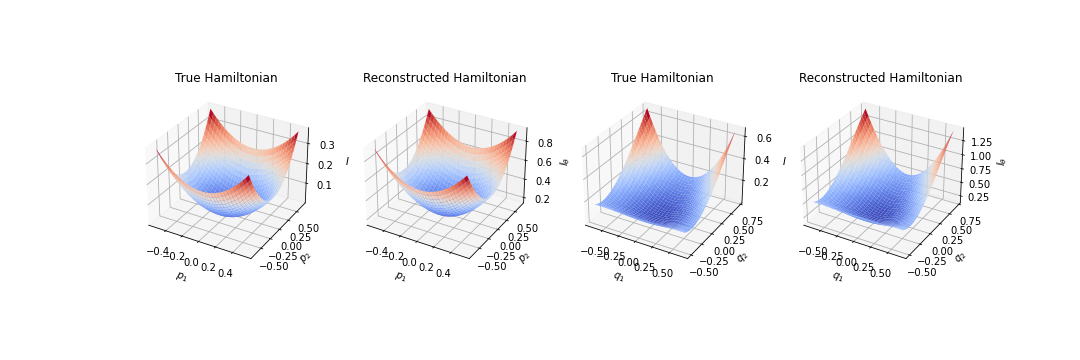}
         \label{sfig:hh_shapes}
     \end{subfigure}
     \caption{Numerical results for the H\'{e}non--Heiles model. \textit{Top row (left):} Number of first integrals identified using the explained ratio.  \textit{Top row (center):} Simulation of a trajectory. \textit{Top row (right):} Relative conservation error.   \textit{Bottom row:} The shape of the learned first integral $I_{\boldsymbol{\theta}}$. First two columns show the four dimensional shape in the first two coordinates and the last two columns in the last two coordinates.}
    \label{fig:hh_nn_conservation}
\end{figure}

Training four neural networks and computing the explained ratio (see Eq.~\eqref{eq:matrix_A}) leads to one found conservation law, see Fig.~\ref{fig:hh_nn_conservation}. The Hamiltonian is a four dimensional function and we visualize different parts of it in Fig.~\ref{fig:hh_nn_conservation}. We see that each part is well resembled by the learned quantity and that the learned conservation law $I_{\boldsymbol{\theta}}$ is conserved in the order of $\mathcal{O}(10^{-2})$. Moreover, when simulating a trajectory using $I_{\boldsymbol{\theta}}$ and $\bar\epsilon_{\boldsymbol{\theta}}$, we see that the trajectory computed using the neural network is close to the numerical reference simulation. With this example we demonstrate that even for a high order system with only one conserved quantity our method correctly finds exactly one conservation law.

\subsection{Conservative Lorenz--1963 system}

As our next example, we consider the conservative Lorenz--1963 system given by
\begin{align}\label{eq:lorenz}
\begin{split}
    \dot{x}=\sigma y, \quad
    \dot{y}=x(\rho-z),\quad
    \dot{z}=xy,
\end{split}
\end{align}
which has the two first integrals
\begin{align*}
    I^1 &= z-\frac{x^2}{2\sigma},\quad
    I^2 = \frac{y^2}{2}+\frac{z^2}{2}-\rho z,
\end{align*}
see~\cite{nevi94a} for further discussions. It was shown in~\cite{nevi94a} that one can rewrite system \eqref{eq:lorenz} as
\begin{equation}\label{eq:LorenzSkewGradientForm}
    \dot{\mathbf{x}} = \epsilon \nabla I^1\nabla I^2
\end{equation}
where $\mathbf{x}=(x,y,z)$ and $\epsilon$ is the constant totally anti-symmetric $3\times 3 \times 3$ Levi-Civita tensor.

There are three degrees of freedom for the conservative Lorenz--1963 model, and accordingly three neural networks have to be trained. With the conservative Lorenz--1963 model possessing two first integrals, all three methods presented in Section~\ref{sec:Method} could be showcased. 

\begin{figure}[!ht]
\begin{subfigure}[b]{0.24\textwidth}
    \centering
    \includegraphics[width=\textwidth, height=25mm]{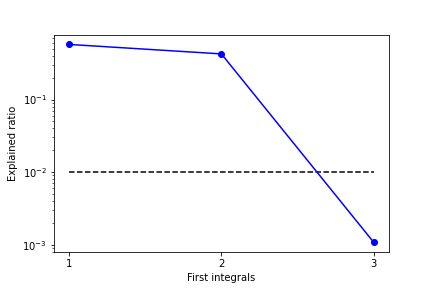}
    \label{fig:LZRank}
\end{subfigure}
\begin{subfigure}[b]{0.24\textwidth}
    \centering
    \includegraphics[width=\textwidth, height=32mm]{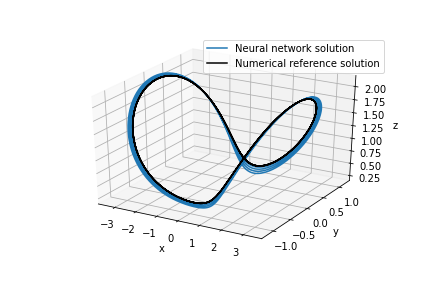}
\end{subfigure}
\begin{subfigure}[b]{0.48\textwidth}
    \centering
    \includegraphics[width=\textwidth, height=30mm]{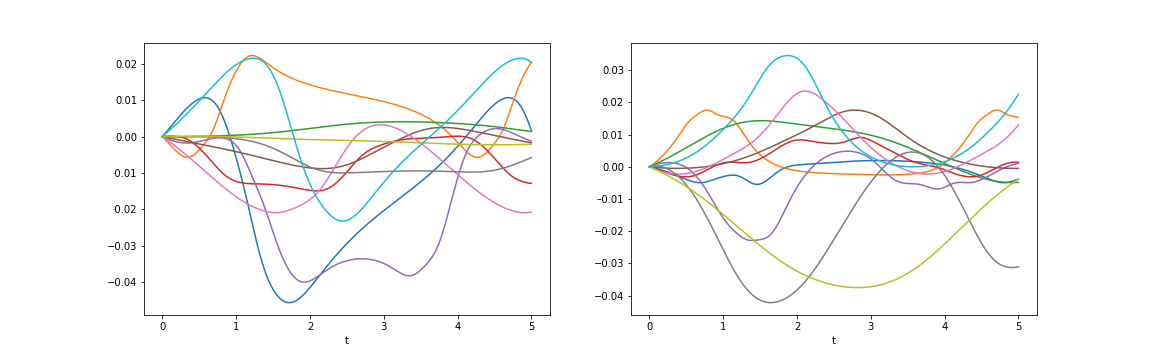}
\end{subfigure}
\caption{Numerical results for the conservative Lorenz--1963 model. \textit{First plot:} Number of first integrals identified using the explained ratio. \textit{Second plot:} Using the learned first integral $I_{\boldsymbol{\theta}_1}$ and the associated anti-symmetric matrix $\bar\epsilon_{\boldsymbol{\theta}_1}$ we can simulate a trajectory for the Lorenz--1963 model. Excellent numerical agreement with a numerical reference solution is observed. \textit{Third plot:} Relative conservation error for $I_{\boldsymbol{\theta}_1}$ along 10 randomly generated trajectories. \textit{Fourth plot:} Relative conservation error for $I_{\boldsymbol{\theta}_2}$ along 10 randomly generated trajectories.}
\label{fig:NumericalResultsLorenz}
\end{figure}

The results of the three trained neural networks are depicted in Fig. \ref{fig:NumericalResultsLorenz}. Notably, this figure contains the computation of the numerical rank, via the explained ratio, of matrix~$A$ from Eq.~\eqref{eq:matrix_A}, which verifies that indeed two functionally independent first integrals have been found.

Once the first integrals and associated anti-symmetric matrices have been found, they can be used for a simulation of new trajectories, verifying that the correct dynamics of the conservative Lorenz--1963 model were understood by the neural networks. Crucially, since using the penalty-based method we did not learn the full skew-gradient form of the conservative Lorenz--1963 model~\eqref{eq:LorenzSkewGradientForm} but rather the reduced forms described in Eq.~\eqref{eq:SelfConsistencyLossPenaltyForm}. This means that we can simulate trajectories of the conservative Lorenz--1963 model using either $I_{\boldsymbol{\theta}_1}$ (and the associated anti-symmetric matrix $\bar\epsilon_{\boldsymbol{\theta}_1}$) or $I_{\boldsymbol{\theta}_2}$ (and the associated anti-symmetric matrix $\bar\epsilon_{\boldsymbol{\theta}_2}$). Fig.~\ref{fig:NumericalResultsLorenz} depicts the trajectory simulations using $I_{\boldsymbol{\theta}_1}$. The results using $I_{\boldsymbol{\theta}_2}$ are qualitatively comparable and are thus not shown here. Lastly, Fig.~\ref{fig:NumericalResultsLorenz} also shows the relative conservation error of $I_{\boldsymbol{\theta}_1}$ and $I_{\boldsymbol{\theta}_2}$ along 10 randomly generated trajectories for the conservative Lorenz--1963 model. These show that both of the learned first integrals are conserved to order $\mathcal O(10^{-2})$.

\begin{figure}[!ht]
    \centering
    \begin{subfigure}[b]{0.24\textwidth}
         \centering
         \includegraphics[width=\textwidth]{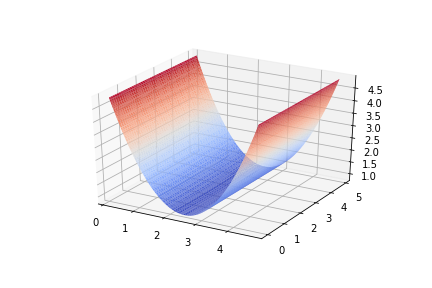}
     \end{subfigure}
    \begin{subfigure}[b]{0.24\textwidth}
         \centering
         \includegraphics[width=\textwidth]{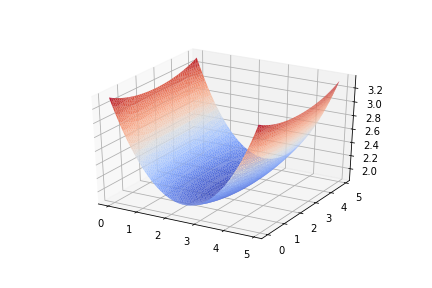}
     \end{subfigure}
     \begin{subfigure}[b]{0.24\textwidth}
         \centering
         \includegraphics[width=\textwidth]{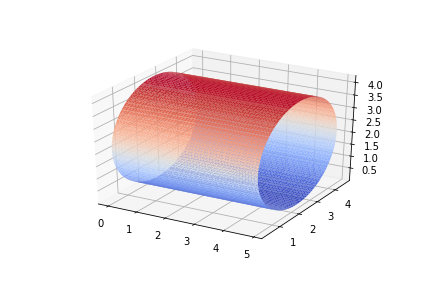}
     \end{subfigure}
    \begin{subfigure}[b]{0.24\textwidth}
         \centering
         \includegraphics[width=\textwidth]{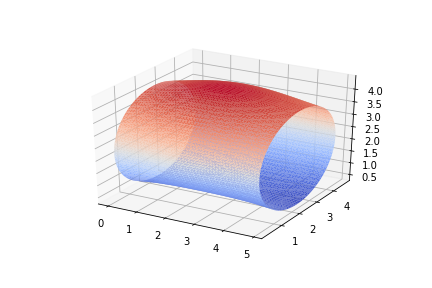}
     \end{subfigure}    
    \caption{The shape of the two first integrals. \textit{First plot:} Learned first integral $I_{\boldsymbol{\theta}_1}$. \textit{Second plot:} True first integral $I^1$. \textit{Third plot:} Learned second first integral $I_{\boldsymbol{\theta}_2}$. \textit{Fourth plot:} True second first integral $I^2$.}
    \label{fig:Lorenz_I_shape}
\end{figure}

Figure~\ref{fig:Lorenz_I_shape} finally shows the shapes of the two learned first integrals $I_{\boldsymbol{\theta}_1}$ and $I_{\boldsymbol{\theta}_2}$, both of which closely resemble those of the true first integrals $I^1$ and $I^2$.

\subsection{Three-species Lotka--Volterra equations}

\begin{figure}[!ht]
    \centering
    \begin{subfigure}[b]{0.24\textwidth}
         \centering
         \includegraphics[width=\textwidth]{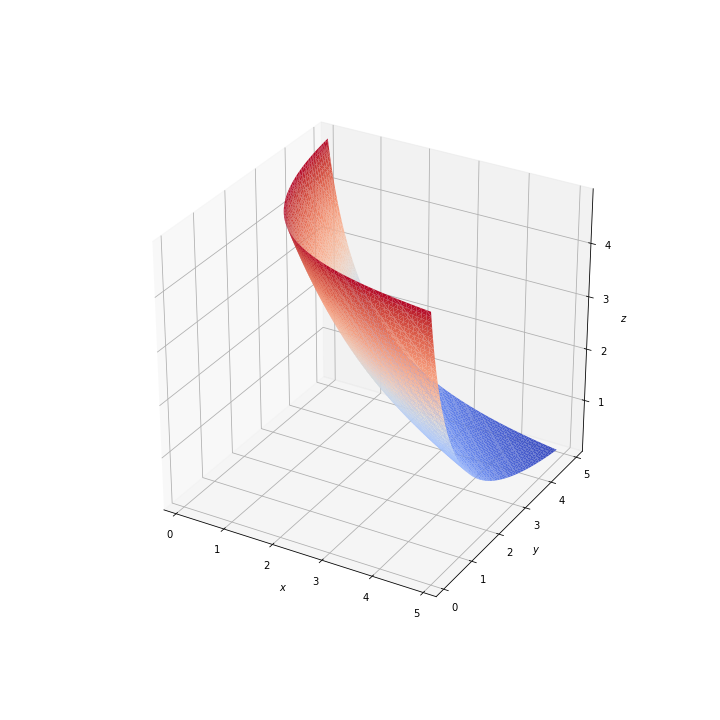}
     \end{subfigure}
    \begin{subfigure}[b]{0.24\textwidth}
         \centering
         \includegraphics[width=\textwidth]{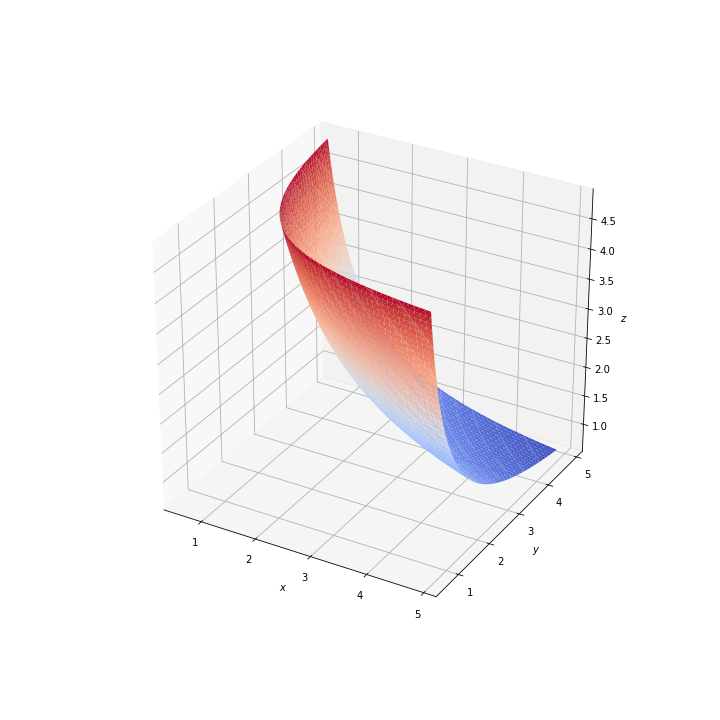}
     \end{subfigure}
     \begin{subfigure}[b]{0.24\textwidth}
         \centering
         \includegraphics[width=\textwidth]{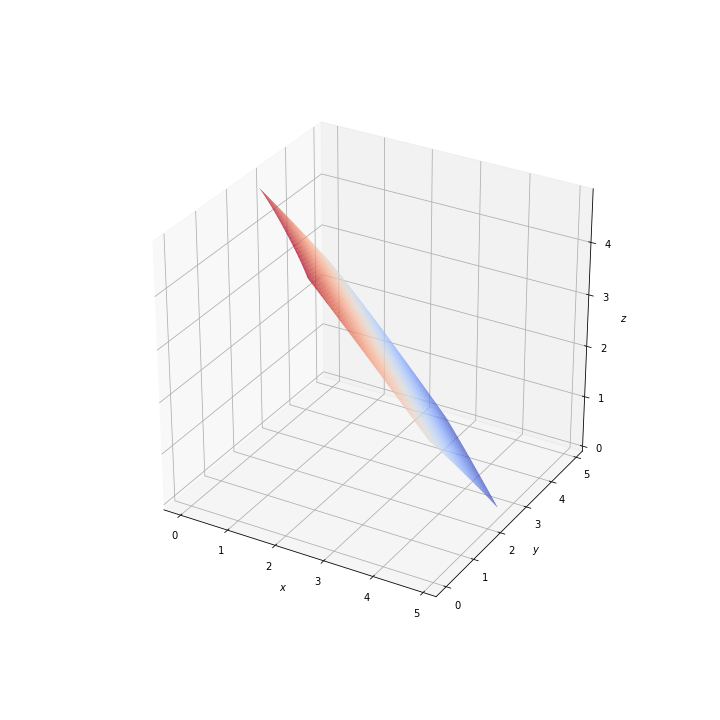}
     \end{subfigure}
    \begin{subfigure}[b]{0.24\textwidth}
         \centering
         \includegraphics[width=\textwidth]{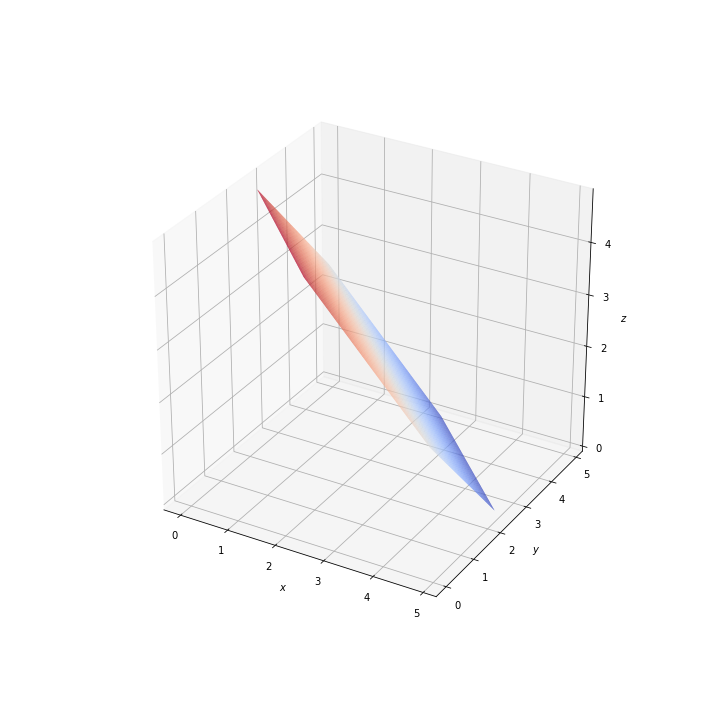}
     \end{subfigure}    
    \caption{The shape of the two first integrals. \textit{First plot:} Learned first integral $I_{\boldsymbol{\theta}_1}$. \textit{Second plot:} True first integral $I^1$. \textit{Third plot:} Learned second first integral $I_{\boldsymbol{\theta}_2}$. \textit{Fourth plot:} True second first integral $I^2$.}
    \label{fig:LK_I_shape}
\end{figure}

We consider the three-species Lotka--Volterra system
\begin{align}\label{eq:LotkaVolterraModel}
\begin{split}
 \dot x = x(y-z),\quad
 \dot y = y(z-x),\quad
 \dot z = z(x-y),
\end{split}
\end{align}
which has two first integrals
\[
I^1 = x + y + z,\quad I^2 = xyz,
\]
see Fig.~\ref{fig:LK_I_shape} and refer to~\cite{schi03a} for further details. The skew-gradient form for the three-species Lotka--Volterra model coincides with Eq.~\eqref{eq:LorenzSkewGradientForm}. We include a discussion of the Lotka--Volterra equations here, as in comparison with the conservative Lorenz--1963 model it admits a cubic first integral, and standard geometric numerical integration methods traditionally struggle with such integrals~\cite{haie06a}.

\begin{figure}[!ht]
\begin{subfigure}[b]{0.24\textwidth}
    \centering
    \includegraphics[width=\textwidth, height=25mm]{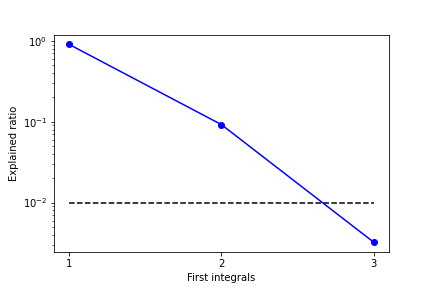}
    \label{fig:LKRank}
\end{subfigure}
\begin{subfigure}[b]{0.24\textwidth}
    \centering
    \includegraphics[width=\textwidth, height=30mm]{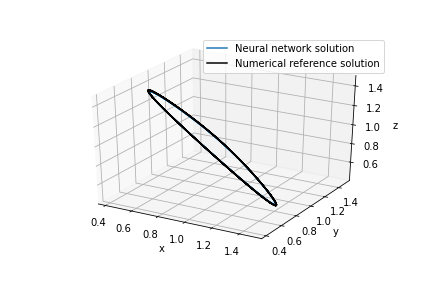}
\end{subfigure}
\begin{subfigure}[b]{0.24\textwidth}
    \centering
    \includegraphics[width=\textwidth, height=30mm]{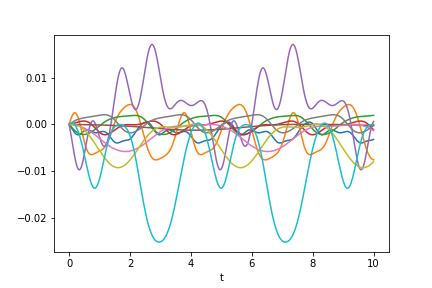}
\end{subfigure}
\begin{subfigure}[b]{0.24\textwidth}
    \centering
    \includegraphics[width=\textwidth, height=30mm]{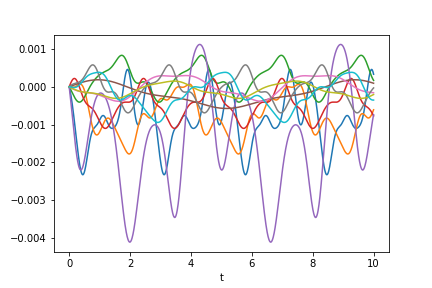}
\end{subfigure}
\caption{Numerical results for the three-dimensional Lotka--Volterra model. \textit{First plot:} Number of first integrals identified using the explained ratio. \textit{Second plot:} Using the learned first integral $I_{\boldsymbol{\theta}_1}$ and the associated anti-symmetric matrix $\bar\epsilon_{\boldsymbol{\theta}_1}$ we can simulate a trajectory for the Lotka--Volterra model. Excellent numerical agreement with a numerical reference solution is observed. \textit{Third and fourth plot:} Relative conservation error for $I_{\boldsymbol{\theta}_1}$ and $I_{\boldsymbol{\theta}_2}$ along 10 randomly generated trajectories.}
\label{fig:NumericalResultsLK}
\end{figure}

The Lotka--Volterra equations have three degrees of freedom and thus three neural networks were trained. We determined the number of functionally independent first integrals by evaluating the explained ratio using the matrix $A$ from Eq.~\eqref{eq:matrix_A} which verifies that there are indeed two first integrals, see Fig.~\ref{fig:NumericalResultsLK}. Once the first integrals have been learned, we can once again used them to simulate new trajectories. Due to using the iterative approach in computing the first integrals and associated anti-symmetric matrices, either $I_{\boldsymbol{\theta}_1}$ or $I_{\boldsymbol{\theta}_2}$, along with the associated matrices $\bar\epsilon_{\boldsymbol{\theta}_1}$ or $\bar\epsilon_{\boldsymbol{\theta}_2}$, could be used. In Fig.~\ref{fig:NumericalResultsLK} we only show the results using the first learned first integral as the results using the second learned first integral are qualitatively similar. Lastly, numerical conservation along trajectories for the learned first integrals is also verified in Fig.~\ref{fig:NumericalResultsLK}, with the numerical conservation error being of order $\mathcal{O}(10^{-2})$.

Figure~\ref{fig:LK_I_shape} the shows the shapes of the two learned first integrals, both of which closely resemble those of the true first integrals $I^1$ and $I^2$.

\subsection{Point vortex dynamics}

The equations for the $p$-point vortex problem in the plane are given by
\begin{align}\label{eq:point_vortex}
\begin{split}
\dot x_i = -\frac{1}{2\pi}\sum_{j=1,j\ne i}^p\Gamma_j\frac{y_{ij}}{r_{ij}^2},\quad
\dot y_i = \frac{1}{2\pi}\sum_{j=1,j\ne i}^p\Gamma_j\frac{x_{ij}}{r_{ij}^2},
\end{split}
\end{align}
where $i=1,\dots,p$, and $(x_i,y_i)$ denotes the position of the $i$th vortex with vortex strength $\Gamma_i$, and $x_{ij}=x_i-x_j$, $y_{ij}=y_i-y_j$ and $r_{ij}=\sqrt{x_{ij}^2+y_{ij}^2}$. This system admits the following four first integrals,
\begin{align}\label{eq:PointVortexFirstIntegrals}
\begin{split}
I^1 &= \sum_{i=1}^p\Gamma_ix_i,\quad
I^2 = \sum_{i=1}^p\Gamma_iy_i,\\
I^3 &= \sum_{i=1}^p\Gamma_i(x_i^2+y_i^2),\quad
I^4 = -\frac{1}{2\pi}\sum_{1\leqslant i< j\leqslant p}\Gamma_i\Gamma_j\log r_{ij},
\end{split}
\end{align}
signifying planar momentum in $x$- and $y$-direction, angular momentum, and energy, respectively. For a more in-depth description of this model, consult~\cite{aref83a,mulle14a}.

Let $\mathbf{z}=(z_1,z_2,z_3,z_4)=(x_1, x_2, y_1, y_2)$. The skew-gradient form for system~\eqref{eq:point_vortex} reads
\[
\dot z_i = \tilde\epsilon_{ijklm}\partial_jI^1\partial_kI^2 \partial_lI^3\partial_mI^4,
\]
for some anti-symmetric 5-tensor $\tilde\epsilon$. Since we are solving the problem in an iterative fashion using the third method proposed in Section~\ref{sec:Method} we re-cast this problem using the simpler skew-matrix form. Let $\epsilon^k = (\epsilon_{ij})^k$ be an anti-symmetric $4\times 4$ matrix for $k=1\dots, 4$, using any of the four first integrals, we can then re-write system~\eqref{eq:point_vortex} in the form
\begin{equation}
    \dot z_i =(\epsilon_{ij}) ^k\partial_j I^k,\qquad \textup{(no summation over $k$)}
\end{equation}
and the goal of the neural networks is to identify both $\epsilon^k$ and $I^k$ for all $k=1,\dots,4$.

\begin{figure}[!ht]
\centering
    \begin{subfigure}[b]{0.75\textwidth}
         \centering
         \includegraphics[width=0.49\textwidth]{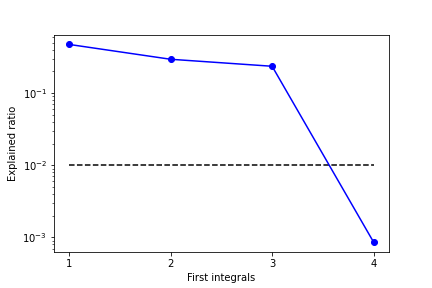}
         \includegraphics[width=0.49\textwidth]{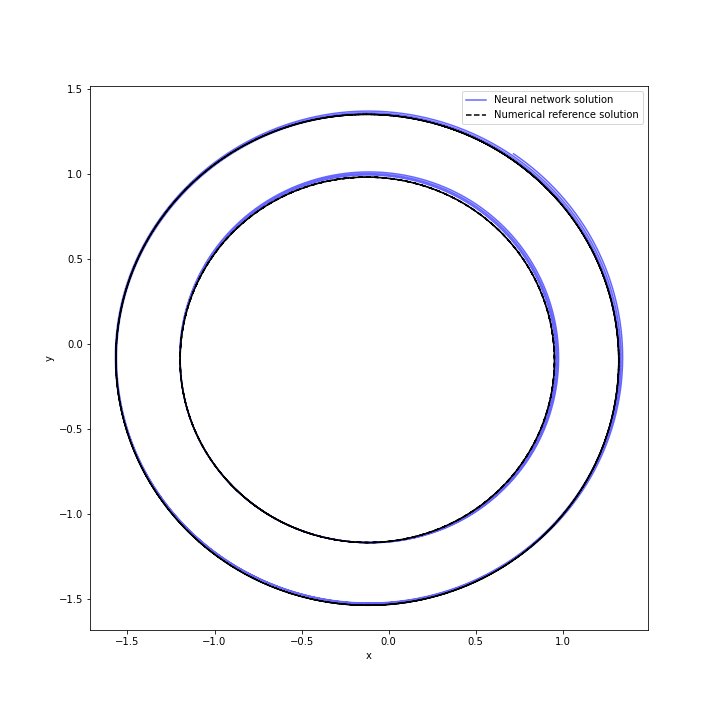}
     \end{subfigure}
     
    \begin{subfigure}[b]{\textwidth}
         \centering
         \includegraphics[width=\textwidth]{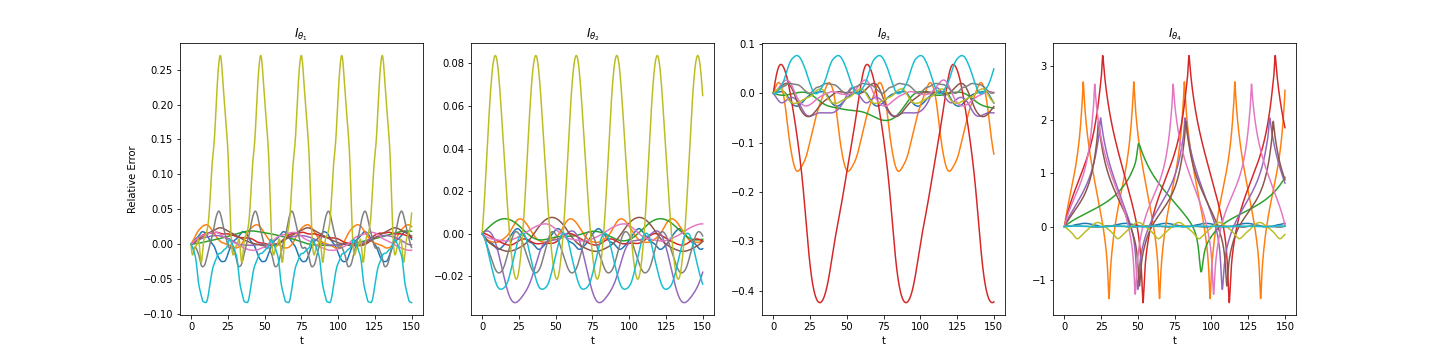}

     \end{subfigure}

     \caption{Evaluation of the learned quantities for the Point--Vortex. \textit{Top left:} Number of first integrals identified using the explained ratio. \textit{Top right:} Using the learned first integral $I_{\boldsymbol{\theta}_1}$ and the associated anti-symmetric matrix $\bar\epsilon_{\boldsymbol{\theta}_1}$ we can simulate a trajectory for the Point--Vortex model. Excellent numerical agreement with a numerical reference solution is observed. \textit{Bottom row:} Relative conservation error for $I_{\boldsymbol{\theta}_1},\ldots,I_{\boldsymbol{\theta}_4}$ along ten randomly generated trajectories.
     }
    \label{fig:PV_shape}
\end{figure}

Using the explained ratio, three out of four conservation laws were learned. In Fig.~\ref{fig:PV_shape} we can see that for the first three learned conservation laws the conservation is in the order of $\mathcal{O}(10^{-1})$ while the learned quantity rejected as a conservation law is not conserved.  



\section{Conclusions}\label{sec:Conclusions}

We have proposed a novel method for identifying first integrals from given trajectory data using neural networks. Our method does not require any additional information besides the given data, and in particular does not require knowledge of an underlying system of differential equations. As a byproduct, once the first integrals have been learned, also the underlying system of differential equations will have been learned, allowing one to use the trained neural networks to generate novel trajectory data. We have demonstrated this method for several models of mathematical physics.

While carrying out our experiments we have demonstrated the robustness of our method by generating trajectory data that only approximately preserves the first integrals of the associated systems of differential equations. We believe this robustness is critical as real-world data is never without noise and any meaningful machine learning algorithm has to be able to tolerate some level of corruption of the input data. We have demonstrated this robustness here by considering data that preserves first integrals up to both some small-scale oscillatory behaviour, as inherent in the symplectic Euler methods for Hamiltonian systems, and larger-scale drift, as the exhibited by the trapezoidal rule for quadratic first integrals. Should the given trajectory data include even larger levels of noise then we suggest that the model to learn should be split into a noise-free part, given by the discrete gradient form, and a dedicated noise model as suggested in, e.g.~\cite{rudy19a}. We will reserve a dedicated study of this case for future research.

\section*{Acknowledgements}

The research of SA and AB was supported by the Canada Research Chairs program and the NSERC Discovery Grant program. The research of RB is funded by the Deutsche Forschungsgemeinschaft (DFG, German Research Foundation) – Project-ID 274762653 – TRR 181. The research of PH was supported by the Specific Research Grant SGS/13/2020 of Silesian University in Opava. Part of this work was done in the course of PH's visit to the Memorial University of Newfoundland, and he thanks Alex Bihlo and his group for the warm hospitality extended to him.

\bibliography{mybib}
\end{document}